\documentclass{article}
\usepackage{spconf,amsmath,graphicx}
\usepackage{amsfonts}
\usepackage{bbm}
\usepackage{color}
\usepackage{times}
\usepackage{soul}
\usepackage{url}
\usepackage[utf8]{inputenc}
\usepackage{graphicx}
\usepackage{amsmath}
\usepackage{amsthm}
\usepackage{booktabs}
\usepackage{algorithm}
\usepackage{algorithmic}
\usepackage{amsthm,amssymb,mathrsfs}
\usepackage{multirow}
\usepackage{stfloats}
\usepackage{bm}
\usepackage{changes}
\usepackage{makecell}
\usepackage{balance}
\title{Improving Biomedical Entity Linking with Retrieval-enhanced Learning}
\name{Zhenxi Lin, Ziheng Zhang, Xian Wu, Yefeng Zheng}
\address{Jarvis Research Center, Tencent YouTu Lab, Shenzhen, China}

\begin{document}
%
\maketitle
\begin{abstract}
Biomedical entity linking (BioEL) has achieved remarkable progress with the help of pre-trained language models.
However, existing BioEL methods usually struggle to handle rare and difficult entities due to long-tailed distribution.
To address this limitation, we introduce a new scheme $k$NN-BioEL, which provides a BioEL model with the ability to reference similar instances from the entire training corpus as clues for prediction, thus improving the generalization capabilities.
Moreover, we design a contrastive learning objective with dynamic hard negative sampling (DHNS) that improves the quality of the retrieved neighbors during inference.
Extensive experimental results show that $k$NN-BioEL outperforms state-of-the-art baselines on several datasets.\footnote{The source code is available at https://github.com/lzxlin/kNN-BioEL.}
\end{abstract}
\section{Introduction}
\label{sec:intro}

Biomedical entity linking (BioEL) aims at mapping mentions (e.g., diseases and drugs) to standard entities in a biomedical ontology, such as UMLS~\cite{bodenreider2004unified}, and BioEL is essential for various biomedical tasks, including ontology construction~\cite{xiang2021ontoea}, automatic diagnosis~\cite{yuan2021efficient}, and drug-drug interaction prediction~\cite{zhang2023emerging}.
Unlike entity linking in the general domain, BioEL is characterized by diverse and fine-grained entity names.
Firstly, biomedical entities can have various names, including synonyms and morphological variations.
For instance, the entity \textit{lung cancer} is also denoted as \textit{lung neoplasm malignant}. Moreover, biomedical entities sharing similar surface forms could possess distinct meanings and cause mismatches, such as \textit{Type 1 Diabetes} and \textit{Type 2 Diabetes}.

The early attempts relied on string-matching~\cite{leaman2016taggerone} or supervised multi-class classifiers~\cite{niu2019multi}, which were limited to matching morphologically similar terms or handling large-scale ontologies.
With the development of contextualized representations, the dominant methods~\cite{sung2020biomedical, lai2021bert, liu2021self, wang2022prompt} in BioEL adopted pre-trained language models and fine-tuned them to effectively capture the semantic information of entities.
These methods primarily utilized bi-encoder models to encode mentions and entities into the same embedding space and then linked them using embedding similarities.
Some previous studies~\cite{xu2020generate, zhu2021enhancing, xu2023improving} further applied a cross-encoder to boost performance by capturing fine-grained mention-entity interactions on top of a bi-encoder.
Another type of BioEL involved the use of generative models~\cite{yuan2022generative, yuan2022biobart} that directly generated linked entities, thereby circumventing the need for negative sample mining.

\begin{figure*}[!ht]
\begin{minipage}[t]{0.41\textwidth}
    \centering
    \includegraphics[width=0.95\linewidth]{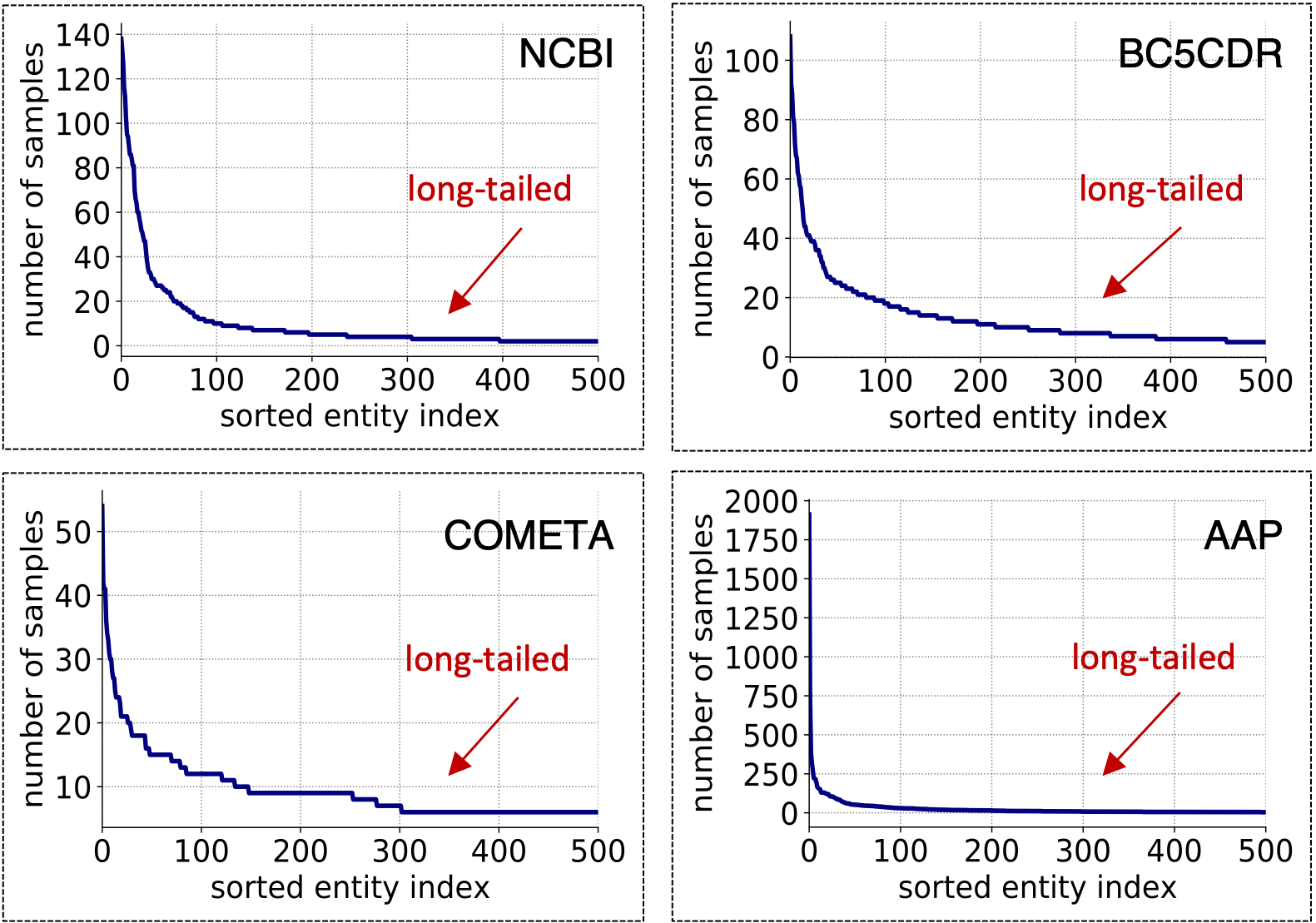}
    \caption{The long-tailed distribution of medical entities in four BioEL datasets.}
    \label{fig:dist}
\end{minipage}
\begin{minipage}[t]{0.57\textwidth}
    \centering
    \includegraphics[width=0.95\linewidth]{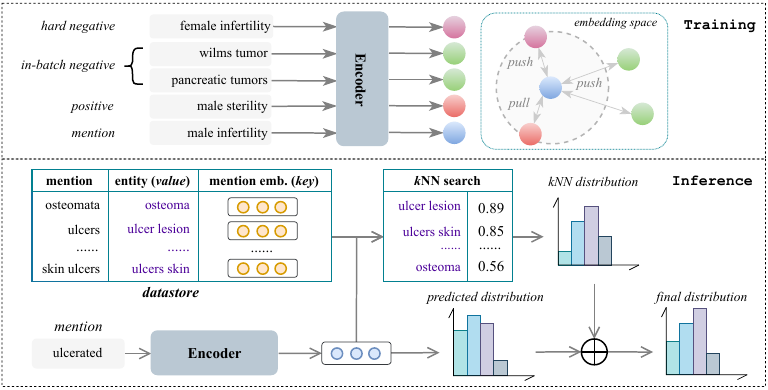}
    \caption{The overview of $k$NN-BioEL training process (\textit{above}) and $k$NN-BioEL inference process (\textit{below}).}
    \label{fig:model}
\end{minipage}
\end{figure*}

However, a major challenge for existing methods is the long-tailed entities.
We observe the long-tailed distribution in multiple BioEL datasets due to the large scale of medical ontology and costly expert annotation, as seen in Figure~\ref{fig:dist}.
The semantics of long-tailed entities cannot be fully captured by existing methods. Hence, it is non-trivial to develop models that can effectively generalize and handle long-tailed entities.

Inspired by the recent retrieval-enhanced model in machine translation~\cite{khandelwal2020nearest}, we propose a simple but effective retrieval-enhanced learning paradigm to address BioEL, called \textit{k}NN-BioEL, which utilizes the knowledge from existing similar instances when predicting entity labels.
For example, given a mention \textit{back molar}, a BioEL model could misjudge it as \textit{structure of first molar tooth}, but with reference to a similar instance (\textit{molar}, \textit{structure of molar tooth}), the model could output the correct prediction \textit{structure of molar tooth} (ID:76928009).
In this work, we construct a datastore that retains the mention embeddings generated by a vanilla BioEL model, along with their corresponding entity labels. During inference, the model retrieves the top-$k$ nearest neighboring instances as clues from the datastore based on mention embedding similarity and then makes inferences by linearly interpolating the model predicted distribution with their nearest neighbor distribution.
Furthermore, to accurately capture fine-grained semantic differences between different biomedical entities (e.g., \textit{Type 1 Diabetes} and \textit{Type 2 Diabetes}) and improve the soundness of retrieved neighbors, we incorporate a contrastive learning (CL) objective with dynamic hard negative sampling (DHNS) during training.
It is worth noting that our method is highly versatile and can be directly applied to most existing BioEL models.
We experimentally validate the effectiveness and superiority of \textit{k}NN-BioEL as it achieves state-of-the-art performance on several benchmark datasets.
The model analysis results also suggest that \textit{k}NN-BioEL is capable of dealing with long-tailed biomedical entities, thus improving the generalization capabilities.

\section{Method}
\label{sec:method}

We start with the problem formulation and the notations. Formally, given a pre-identified textual mention $m$ and an ontology $\mathcal{E}$ consisting of $N$ candidate entities, the goal of BioEL is to map the mention $m$ to its correct entity $e\in\mathcal{E}$.
For simplicity, we do not consider the use of any auxiliary information, such as mention context~\cite{zhu2021enhancing} or entity descriptions~\cite{xu2023improving}.
The overall architecture of the proposed \textit{k}NN-BioEL is shown in Figure~\ref{fig:model}.
In the subsequent sections, we provide details on how to incorporate retrieval-enhanced learning into both the training and inference processes of the BioEL model.

\subsection{BioEL Model}
\label{ssec:model}

Most BioEL models adopt a two-stage method to find the correct entity link for a given mention due to its prominent performance. A bi-encoder is utilized in the first stage to independently embed the mention and entity name and then generate top candidates. Then a cross-encoder is used to re-rank each candidate by concatenating the mention and entity name. The drawback of cross-encoder is its higher computational complexity and inference speed compared to bi-encoder.
To showcase the flexibility and practicality of our approach, we exclusively employ the bi-encoder as our BioEL model.

Following the previous work~\cite{zhu2021enhancing,xu2023improving}, we employ SapBERT~\cite{liu2021self} as the shared encoder to generate dense vectors for both mentions and entities. 
The mention embedding $f(m)$ of mention $m$ can be formulated as:
\begin{gather}
    f(m) = \texttt{SapBERT}(m)\texttt{[CLS]}\label{eq:emb_m},
\end{gather}
where $\texttt{[CLS]}$ denotes that the final hidden state of the special token is utilized to derive a fixed-size feature vector.
The entity embedding $f(e)$ of entity $e$ is computed similarly.
The score of a mention-entity pair $(m,e)$ is denoted as follow:
\begin{gather}
    s(m, e) = g(f(m), f(e)),
\end{gather}
where $g$ is the cosine similarity.

\subsection{Model Training}
\label{ssec:train}

\noindent\textbf{Contrastive Learning (CL)}.
The BioEL model is trained to learn effective representations through contrastive learning~\cite{gao2021simcse}, which aims to optimize the agreement between true mention-entity pairs and false pairs. 
During training, in-batch negative sampling is employed to generate false pairs, where each mention is paired with multiple negative entities within the same batch. However, in-batch negative sampling is inherently limited by the mini-batch size, making it challenging to capture fine-grained semantic differences between entities.
Typically, there are several different surface forms for biomedical entities. On one hand, these surface forms can exhibit significant variations. For example, two surface forms, \textit{motrin} and \textit{ibuprofen} refer to the same chemical entity. On the other hand, surface forms with high similarity can also have different meanings, for instances, disease entities like \textit{Type 1 Diabetes} and \textit{Type 2 Diabetes}, or chemical entities like \textit{xyloglucan endotransglycosylase} and \textit{xyloglucan endoglucanase}. Hence, We further incorporate dynamic hard negative sampling (DHNS) to find the more informative negative examples for accurately capturing fine-grained semantic relationships in biomedical entities.

\noindent\textbf{Dynamic Hard Negative Sampling (DHNS)}.
Instead of random negatives or in-batch negatives, DHNS constructs global negatives using the being-optimized BioEL model to retrieve from all the entities.
DHNS pre-computes all entity embeddings using the encoder and constructs the entity index. 
In each epoch, DHNS leverages the mention embeddings within the batch to retrieve the top false entities, serving as hard negatives. These hard negatives are dynamically updated to include more challenging negative samples for each epoch.

\noindent\textbf{Optimization Objective}.
For each training pair $(m, e)$ within the batch $\mathcal{B}$, where entity $e$ is the corresponding label of mention $m$. 
The negative includes two types: $e'\in E_1$, obtained from in-batch negative sampling, and $e''\in E_2$, obtained from DHNS. Specifically, $E_1\subseteq\mathcal{E}$ represents the entity set of other mentions within $\mathcal{B}$, while $E_2$ represents the top-$p$ most similar but incorrect entities for $m$ retrieved from the entity set $\mathcal{E}$.
The objective can be formulated as:
\begin{equation}
\begin{small}
\begin{aligned}
    \mathcal{L}(m,e)=
    -\log
    \left( 
    \frac{\delta(m,e)}
    {\delta(m,e)+ \sum\limits_{e'\in{E}_1}\delta(m, e') + \sum\limits_{e''\in{E}_2}\delta(m,e'')} 
    \right),
    \label{eq:loss}
\end{aligned}
\end{small}
\end{equation}
where $\delta(m,e)=\exp(s(m,e)/\tau)$ and $\tau$ is a temperature hyper-parameter.
This function encourages the model to pull positive pairs closer together and push negative pairs apart, thereby enhancing the quality of text representations.

\subsection{Model Inference}
\label{ssec:inference}

To improve the generalization capabilities of the BioEL model, we propose a $k$ nearest neighbor retrieval mechanism that leverages existing instances to provide valuable clues during inference. This mechanism consists of two steps: constructing a datastore of training instances and interpolating the model's prediction with the $k$NN distribution.

\noindent\textbf{Datastore Construction}. 
Given the $i$-th instance $(m_i,e_i)$ in the training set $\mathcal{D}$, we define the key-value pair $(f(m_i), e_i)$, in which $f(m_i)$ is the mention embedding computed in Eq.~(\ref{eq:emb_m}), and $e_i$ is the entity label of mention $m_i$. The datastore $(\mathcal{K}, \mathcal{V})$ is thus the set of all key-value pairs constructed from all the training instances in $\mathcal{D}$, i.e., $(\mathcal{K}, \mathcal{V}) = \{(f(m_i),e_i)|(m_i, e_i)\in\mathcal{D}\}$.

\noindent\textbf{$k$NN Distribution Interpolation}. 
Given a mention $x$, the model outputs its mention embedding $f(x)$ via Eq.~(\ref{eq:emb_m}) and generates the predicted distribution over all entities $p_{\text{BioEL}}(y|x)$ as follow:
\begin{gather}
    p_{\text{BioEL}}(y|x) \propto \mathbb{I}_{y=e}\exp(s(x, e) / \beta_1),
\end{gather}
where $\beta_1$ is a temperature hyper-parameter to flatten or sharpen the distribution. 
The model then queries the datastore with $f(x)$ to retrieve its $k$ nearest neighbors $\mathcal{N}$ according to a similarity measure $g(\cdot, \cdot)$, where $g(\cdot, \cdot)$ is the cosine similarity in our experiments.
Then, we apply the softmax function over the similarity scores to compute the distribution over neighbors and aggregate probability mass to form the $k$NN distribution:\footnote{Unlike~\cite{khandelwal2020nearest}, we empirically find that maximum aggregation is better than sum aggregation for BioEL task.}
\begin{equation}
\begin{small}
\begin{aligned}
    p_{\text{kNN}}(y|x) \propto 
    \max_{(f(m_i), e_i)\in\mathcal{N}}\mathbb{I}_{y=e_i}
    \exp(g(f(x), f(m_i)) / \beta_2),
\end{aligned}
\end{small}
\end{equation}
where $\beta_2$ is also a temperature hyper-parameter. 
Note that, we set zero probability for the entities that do not appear in the retrieved targets.
Finally, we interpolate the $k$NN distribution $p_{\text{kNN}}(y|x)$ with the predicted distribution $p_{\text{BioEL}}(y|x)$ using a hyper-parameter $\lambda$ to product the final distribution:
\begin{gather}
    p(y|x) = \lambda p_{\text{kNN}}(y|x) + (1-\lambda) p_{\text{BioEL}}(y|x).
\end{gather}
At inference time, the entity embedding for all candidates can be pre-computed and cached to reduce computational cost.

\section{Experiments}
\label{sec:exp}

\subsection{Experimental Setup}
\label{ssec:setup}

\noindent\textbf{Datasets and Evaluation}. 
Four BioEL datasets are adopted, including NCBI~\cite{dougan2014ncbi}, BC5CDR~\cite{li2016biocreative}, COMETA~\cite{basaldella2020cometa} and AAP~\cite{limsopatham2016normalising}, which focus on different entity types, including diseases, chemicals, and colloquial terms. The detailed dataset statistics are listed in Table~\ref{tab:dataset}.
Following previous works~\cite{sung2020biomedical,liu2021self,yuan2022generative}, we use the same preprocessing and experimental setting and report Acc@1 and Acc@5 for evaluation.

\begin{table}[h]
    \centering
    \scriptsize
    \renewcommand\arraystretch{1.0}
    \caption{Dataset Statistics.}
    \begin{tabular}{@{}lrrrr@{}}
        \toprule
        & NCBI & BC5CDR & COMETA & AAP  \\
        \midrule
        Target entities $|\mathcal{E}|$ & 14,967 & 268,162 & 350,830 & 1,036 \\
        Train instances & 5,784 & 9,285 & 13,489 & 16,826 \\
        Validation instances & 787 & 9,515 & 2,176 & 1,663 \\
        Test instances & 960 & 9,654 & 4,350 & 1,712 \\
        \bottomrule
    \end{tabular}
    \label{tab:dataset}
\end{table}


\noindent\textbf{Implementation Details}.
The number of training epochs is 20 with early-stopping and the batch size is 128.
The number of hard negatives $p$ is 4.
The hyper-parameters $\tau$ and $\lambda$ are set to 0.01 and 0.1, respectively.
Other hyper-parameters are selected via grid search. For NCBI, BC5CDR, COMETA, and AAP, the temperatures $\beta_1$/$\beta_2$ are set to 0.01/1.0, 0.05/5.0, 0.2/1.0, 1.0/1.0, the number of neighbors $k$ to 4, 4, 128, 128,\footnote{The varying hyper-parameter $k$ could be attributed to different scales of medical ontologies or varying noise tolerance in different datasets.} and the learning rate to $1\times 10^{-4}$, $8\times 10^{-5}$, $4\times 10^{-5}$, $4\times 10^{-5}$, respectively.
We use the AdamW optimizer for training the $k$NN-BioEL, and we compute $p(y|x)$ for all candidate entities and return the top nearest candidates for mention $x$ using FAISS,\footnote{https://github.com/facebookresearch/faiss} an efficient similarity search library.

\noindent\textbf{Baselines}. 
We compare the proposed $k$NN-BioEL against previous state-of-the-art BioEL methods, which
can be classified into three categories: 1) \textit{retrieval-based methods} that solely utilize a bi-encoder to retrieval related entities, including BioSyn~\cite{sung2020biomedical}, ResCNN~\cite{lai2021bert} and SapBERT~\cite{liu2021self}; 2) \textit{two-stage methods} that further use a re-rank model to improve performance, including Clustering-based~\cite{angell2021clustering} and Prompt-BioEL~\cite{xu2023improving}; 3) \textit{generative methods} that directly generate linked entities rather than through retrieval, including GenBioEL~\cite{yuan2022generative} and BioBART~\cite{yuan2022biobart}.

\subsection{Overall Results}
\label{ssec:result}

\begin{table*}[ht]
    \centering
    \footnotesize
    \renewcommand\arraystretch{1.0}
    \caption{Comparative experimental results where the best results are in \textbf{bold} with the second best results \underline{underlined}.}
    \begin{tabular}{l|cc|cc|cc|cc@{}}
        \toprule
        \multirow{2}*{Models} & \multicolumn{2}{c|}{NCBI} & \multicolumn{2}{c|}{BC5CDR} & \multicolumn{2}{c|}{COMETA} & \multicolumn{2}{c}{AAP} \\
        & {\scriptsize Acc@1} & {\scriptsize Acc@5} & {\scriptsize Acc@1} & {\scriptsize Acc@5} & {\scriptsize Acc@1} & {\scriptsize Acc@5} & {\scriptsize Acc@1} & {\scriptsize Acc@5} \\
        \midrule
        BioSyn~\cite{sung2020biomedical} & 91.1 & 93.9 & -- & -- & 71.3 & 77.8 & 82.6 & 87.0 \\
        ResCNN~\cite{lai2021bert} & 92.4 & -- & -- & -- & 80.1 & -- & -- & -- \\
        SapBERT~\cite{liu2021self} & 92.3 & 95.5 & -- & -- & 75.1 & 85.5 & 89.0 & \textbf{96.2} \\
        \midrule
        Clustering-based~\cite{angell2021clustering} & -- & -- & 91.3 & -- & -- & -- & -- & -- \\
        Prompt-BioEL~\cite{xu2023improving} & \underline{92.6} & 95.8 & \textbf{93.7} & \textbf{96.6} & \underline{83.7} & \textbf{92.3} & -- & -- \\
        \midrule
        GenBioEL~\cite{yuan2022generative} & 91.9 & \textbf{96.3} & 93.3 & 95.8 & 81.4 & 88.2 & 89.3 & \underline{96.0} \\
        BioBART-base~\cite{yuan2022biobart} & 89.3 & 95.3 & 93.0 & 95.6 & 79.6 & 88.6 & 87.5 & 94.9 \\
        BioBART-large~\cite{yuan2022biobart} & 89.9 & 95.6 & 93.3 & 95.7 & 81.8 & 88.9 & \underline{89.4} & 95.8 \\
        \midrule
        $k$NN-BioEL & \textbf{92.8} & \underline{96.1} & \textbf{93.7} & \underline{96.2} & \textbf{85.7} & \textbf{92.3} & \textbf{90.3} & 95.3 \\
        \quad w/o $k$NN & \textit{92.7} & \textit{96.2} & \textit{92.5} & \textit{95.9} & \textit{81.7} & \textit{90.1} & \textit{89.1} & \textit{95.4} \\
        \quad w/o DHNS & \textit{92.4} & \textit{96.0} & \textit{90.6} & \textit{94.8} & \textit{84.8} & \textit{90.0} & \textit{89.3} & \textit{95.3} \\
        \bottomrule
    \end{tabular}
    \label{tab:overall-perf}
\end{table*}

Table~\ref{tab:overall-perf} reports the performance of $k$NN-BioEL against the previous baselines on four datasets.
Our method surpasses the previous SOTA baseline in Acc@1 on NCBI, COMETA, and AAP, with relative improvements of 0.2\%, 2.4\%, and 1.0\%, respectively, and achieves comparable results in BC5CDR with Prompt-BioEL.
Noteworthily, the competitive methods, GenBioEL and Prompt-BioEL, rely heavily on extra knowledge of synonyms to pre-train the model, but these synonyms may not be available for realistic scenarios, and collecting them is labor-intensive, limiting the scope of its application.
However, $k$NN-BioEL leverages existing training instances to enhance the model without the need for additional knowledge, which demonstrates the effectiveness of the proposed retrieval-enhanced learning.

\subsection{Model Analysis}
\label{ssec:analysis}

\noindent\textbf{Ablation Study}.
To investigate different components in $k$NN-BioEL, we compare $k$NN-BioEL variants without $k$NN distribution interpolation (w/o $k$NN) and without dynamic hard negative sampling (w/o DHNS). 
From the last three lines of Table~\ref{tab:overall-perf}, we observe drops in Acc@1 when removing $k$NN across all datasets except NCBI, which indicates the necessity of $k$NN distribution interpolation.
Besides, DHNS can consistently improve the performance of the base models, attributable to more discriminative text embeddings.

\noindent\textbf{Long-tailed Case Study}.
Table~\ref{tab:case-study} shows two long-tailed entities \textit{ulcer lesion} and \textit{procaine} to illustrate the amendment by $k$NN search.
Without $k$NN, the BioEL model tends to link mentions to morphological similar but incorrect entities, such as linking \textit{ulcerated} to \textit{ulcerated mass}. With $k$NN and the retrieved instance (\textit{ulcers}, \textit{ulcer lesion}), the probability of the correct entity \textit{ulcer lesion} could surpass that of the incorrect entity \textit{ulcerated mass}, which rectifies the prediction.


\begin{table}[h]
    \centering
    \scriptsize
    \renewcommand\arraystretch{1.0}
    \caption{Two test examples from COMETA with predictions made without ($-$) or with ($+$) $k$NN distribution interpolation.}
    \begin{tabular}{@{}l|l|l@{}}
        \toprule
        Mention &  Top-3 Retrieved Instances $(m_i, e_i)$ & Prediction \\
        \midrule
        \multirow{2}*{ulcerated} & \multirow{2}*{\parbox{4.3cm}{
        (ulcers, ulcer lesion), (ulceration, ulcer lesion), (stomach ulcers, stomach ulcer)
        }} & ($-$) ulcerated mass {\color{red} $\times$} \\
        & & ($+$) ulcer lesion {\color{green} $\checkmark$} \\
        \midrule
        \multirow{2}{*}{novacaine} & \multirow{2}{*}{\parbox{4.3cm}{
        (novocaine, procaine), (lidocaine, lignocaine), (gabapentine, gabapentin)
        }} & ($-$) butacaine {\color{red} $\times$} \\
        & & ($+$) procaine {\color{green} $\checkmark$} \\
        \bottomrule
    \end{tabular}
    \label{tab:case-study}
\end{table}

\begin{figure}[!h]
    \centering
    \includegraphics[width=0.95\linewidth]{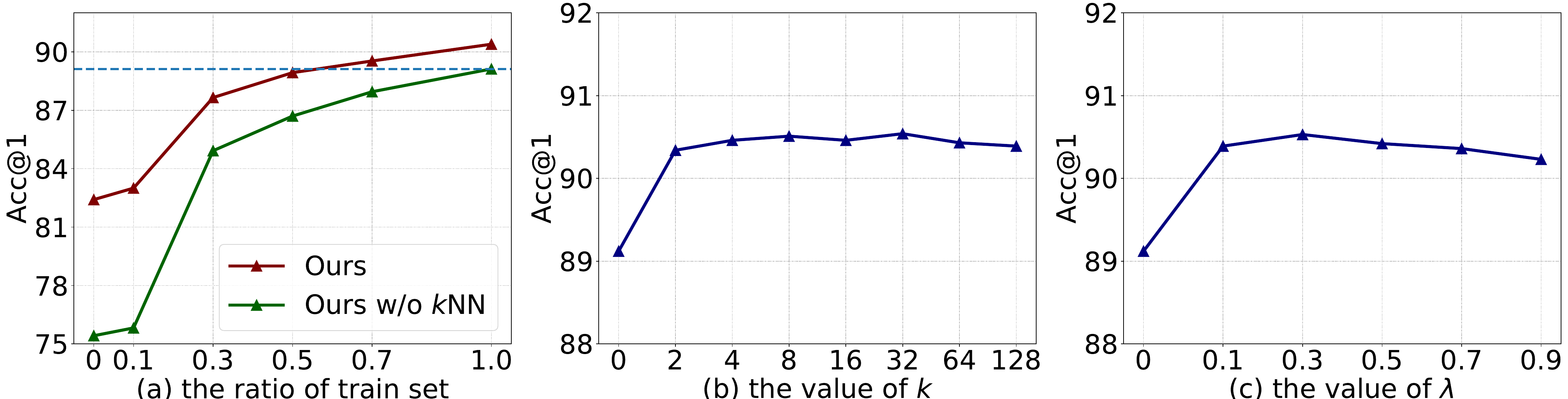}
    \caption{Impact of different hyper-parameters on AAP dataset.}
    \label{fig:analysis}
\end{figure}

\noindent\textbf{Low-resource Scenario Analysis}. 
We further conduct experiments on AAP by varying the percentage of training data used for fine-tuning while holding the full training data in the datastore for $k$NN search.
As shown in Figure~\ref{fig:analysis}(a), even without fine-tuning (i.e., the ratio of train set is 0), our method can significantly boost the vanilla model using $k$NN distribution interpolation.
Moreover, we notice that our method with 50\% training data for fine-tuning can yield comparable performance with the vanilla model trained on full training data (see the blue dotted line).
This result suggests that our method can efficiently improve the generalization performance via the simple retrieval mechanism.

\noindent\textbf{Impacts of Number of Neighbors $k$ and Interpolation Ratio $\lambda$}.
From Figure~\ref{fig:analysis}(b), we find that Acc@1 continues to rise as $k$ increases until convergence, and from Figure~\ref{fig:analysis}(c), with the increase of parameter $\lambda$, Acc@1 increases first and decreases afterward.
These observations illustrate that the ``helpful" instances for queried mentions are limited and excessive instances (larger $k$) or higher weight for instances (larger $\lambda$) could hurt overall performance.

\section{Conclusion}
\label{sec:con}

This paper presented a novel method termed $k$NN-BioEL to address the biomedical entity linking.
kNN-BioEL provides an effective learning paradigm that directly retrieves the knowledge from the existing instances as clues during inference. 
We experimentally validated the state-of-the-art performance of $k$NN-BioEL in several public datasets and its capability of improving model generalization.
For future work, we plan to explore more side information such as mention context or entity descriptions to boost performance.



\vfill\pagebreak

\bibliographystyle{IEEEbib}
\bibliography{strings,refs}

\end{document}